\documentclass[letterpaper, 10 pt, conference]{ieeeconf}
\usepackage{graphicx}
\usepackage{multirow}
\usepackage{colortbl}
\usepackage{hyperref}
\usepackage{stfloats}

\IEEEoverridecommandlockouts                              

\title{\LARGE \bf
Push and Drag: An Active Obstacle Separation Method for Fruit Harvesting Robots
}

\author{Ya Xiong, Yuanyue Ge and P{\aa}l Johan From
\thanks{*This work was supported by Research council of Norway, FORNY2020, project number 2962020.}
\thanks{Authors are with Noronn AS and Faculty of Science and Technology, Norwegian University of Life Sciences,
        {\AA}s, Norway
        {\tt\small \{ya.xiong, yuanyue.ge, pal.johan.from\}@nmbu.no}}%
}

\begin{document}

\maketitle
\thispagestyle{empty}
\pagestyle{empty}

\begin{abstract}
Selectively picking a target fruit surrounded by obstacles is one of the major challenges for fruit harvesting robots. Different from traditional obstacle avoidance methods, this paper presents an active obstacle separation strategy that combines push and drag motions. The separation motion and trajectory are generated based on the 3D visual perception of the obstacle information around the target. A linear push is used to clear the obstacles from the area below the target, while a zig-zag push that contains several linear motions is proposed to push aside more dense obstacles. The zig-zag push can generate multi-directional pushes and the side-to-side motion can break the static contact force between the target and obstacles, thus helping the gripper to receive a target in more complex situations. Moreover, we propose a novel drag operation to address the issue of mis-capturing obstacles located above the target, in which the gripper drags the target to a place with fewer obstacles and then pushes back to move the obstacles aside for further detachment. Furthermore, an image processing pipeline consisting of color thresholding, object detection using deep learning and point cloud operation, is developed to implement the proposed method on a harvesting robot. Field tests show that the proposed method can improve the picking performance substantially. This method helps to enable complex clusters of fruits to be harvested with a higher success rate than conventional methods. 

\end{abstract}

\section{INTRODUCTION}

Fruit production that requires selective harvesting is heavily reliant on human labour \cite{xiong2018design}. This is applicable to crops such as strawberries, sweet peppers, tomatoes, cucumbers, etc. Labour represents the largest cost and also a large operational uncertainty for fruit growers \cite{yamamoto2014development}. Therefore, several attempts have been made to develop a robotic solution for selective harvesting of fruits. Some fruits, such as strawberries and tomatoes, tend to grow in clusters. This makes it difficult to identify and pick individual ripe fruit without damaging or accidentally picking unripe fruit \cite{xiong2019jfr, yamamoto2014development}. Harvesting fruits that grow in clusters or are surrounded by obstacles, such as branches and/or leaves, while leaving the other fruits to remain undamaged on the plant, is one of the primary challenges for fruit harvesting systems \cite{xiong2019jfr, yaguchi2016development}. The surrounding fruits, leaves, stems and other obstacles are often difficult to separate from the target, both in terms of detection and in manipulation.  

In agricultural robotics field, many researchers try to avoid obstacles in both vision and manipulation. To avoid occlusions in sweet pepper picking, a “3D‐move‐to‐see” method was proposed to find the best view with fewer occlusions \cite{lehnert20183d}. To avoid obstacles, a method for cucumber picking was developed that uses a search algorithm to explore the search space for a feasible trajectory, in which each step of the trajectory is checked by a collision detector \cite{van2002autonomous}. Most of the methods found in the literature are passive obstacle avoidance methods, in which the aim is to avoid existing obstacles without changing the environment. However, obstacles are not always avoidable, especially when picking small‐size fruits in clusters, where the obstacles may be extremely close to the targets.

Our previous work presented a gripper for strawberry harvesting that can open fingers to enclose a target from below \cite{xiong2018design}. Without moving the obstacles out of the way, obstacles may prevent the gripper from capturing the target and may also be swallowed with the target if they are located close to the target. Similar problems occur when approaching the fruit from other angles. To solve this issue, in a later work \cite{xiong2019jfr}, we proposed to use a single linear push operation to push aside the obstacles below the target based on the obstacle sensing from a 3D camera. We found that pushing obstacles aside, rather than simply avoiding them makes it possible to pick fruit that would otherwise be inaccessible to the robot. However, a single linear push may be insufficient for dense obstacles from multi-direction with respect to the target, since the linear push moves towards only one direction. Also, the obstacles may be adjacent together that can not be separated during the single push. Furthermore, the gripper may not be able to swallow the whole target but push it up due to the static contact force between the target and obstacles. In addition to that, one frequent failure is the gripper may capture obstacles above the target when it moves up to detach the fruit, which has not been addressed in the previous work. 

In the field of robotic manipulation, most studies focus on obstacle avoidance. Nevertheless, we found some research working on obstacle separation for simple situations. For a warehouse picking application on desk environment, two linear pushing policies were proposed to separate rigid obstacles during the way of the gripper reaching a target bin \cite{danielczuk2018linear}. Another work used Learning from Demonstration (LfD) algorithm for the same application that involves a pushing action \cite{laskey2016robot}. For a similar situation, researchers proposed to use physical engine to calculate the dynamics to predict the object locations for motion planning, which also involves pushing motions \cite{kitaev2015physics, moll2017randomized, dogar2013physics}. Reinforcement learning was also used to train a robot to rearrange objects on a desk using pushing method to make them sparse for individual grasping \cite{zeng2018learning}. However, all these methods were tested at simple environment where some rigid objects were placed on a 2D desk surface. In the agricultural environment, for example strawberry plants, fruits are located in 3D within diverse and unconstrained environments. The flexible peduncles, deformable fruits and many other crop variations make the dynamics difficult to calculate and predict. Moreover, the operation speed of these methods seem very slow, which may not be suitable for fruit harvesting. 

This paper provides the improvements to our previously proposed obstacle separation method \cite{xiong2019jfr}. We extend the pushing policies by adding a zig-zag push and a novel drag operation. We also propose an image processing pipeline to implement the method on a harvesting robot. The proposed method might be also applicable to harvest other fruits, such as tomatoes and cucumbers. 


\section{Methods}
\subsection{Region of Interest} 
\label{RoI}
\begin{figure} [ht]
    \centering
    \includegraphics[width=\linewidth]{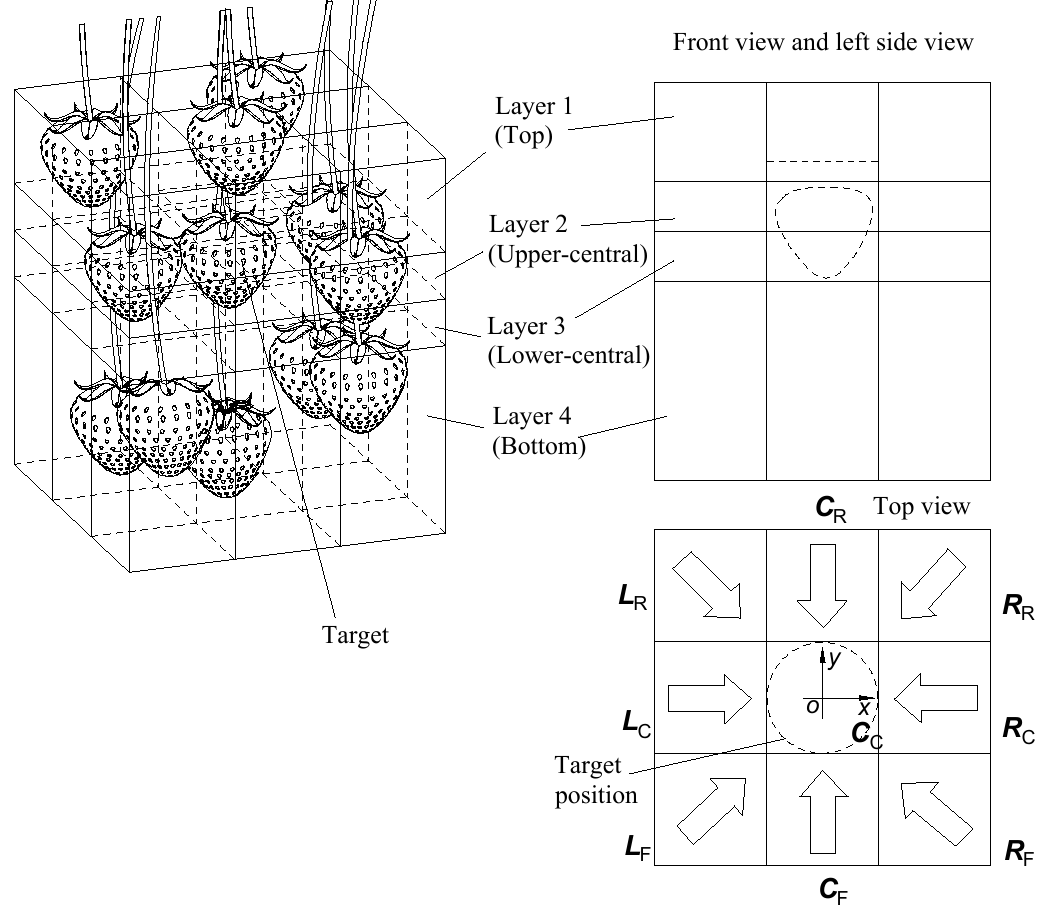}
    \caption{Region of interest area around the target to determine the presence of obstacles.}
    \label{fig:fig2}
\end{figure}
The obstacle separation trajectory is generated according to the visual perception of the obstacle information around the target. We select a region of interest (RoI) area around the target to determine the presence of the obstacles. The RoI comprises a volume of 3D point cloud that contains the target fruit and potentially one or more obstacles. As shown in Fig. \ref{fig:fig2}, the RoI area can be divided into four horizontal layers: a top layer 1, an upper-central layer 2, a lower-central layer 3 and a bottom layer 4. 

As the top view is shown in Fig. \ref{fig:fig2}, each layer of the RoI is further segmented into nine cuboid blocks. On each layer, the blocks are arranged in a 3x3 grid that has its center at the horizontal midpoint of the target strawberry such that the central block $\textit{\textbf{C}}_{\mathrm{C}}$ encompasses the position of the target strawberry in the \textit{xy} plane. 

To generate the separation paths, each block is assigned a horizontal vector representative of the direction from the block to the central block $\textit{\textbf{C}}_{\mathrm{C}}$. The direction of the vector is determined by the position of the block so that all vectors are directed from the center of the corresponding block towards the center of the central block $\textit{\textbf{C}}_{\mathrm{C}}$. We use the number of points $N$ in the block of point cloud to determine whether there are obstacles present in the block or not. 

The gripper is instructed to operate in three distinct stages. As the gripper is picking from below, during the first stage, the gripper moves obstacles horizontally within layer 4. During the second stage, the device moves up to swallow the target and separates the obstacles within layer 2 and layer 3. During the third stage, the gripper drags the target into a picking position with fewer obstacles if the central block $\textit{\textbf{C}}_{\mathrm{C}}$ in layer 1 is occupied. 
The detailed separation policies will be elaborated in the below sections.

\subsection{Horizontal Push} 
The first stage is to separate the obstacles horizontally under the target in layer 4. Compared to our previous method \cite{xiong2019jfr}, we add a zig-zag pushing policy in addition to the single push. A single push means that the gripper linearly pushes the obstacles aside once, starting from the region with fewer obstacles. A zig-zag push is a motion where the gripper uses a zig-zag movement that contains several linear motions to push the obstacles side to side. This motion can not only move the obstacles out of the way, but also break the static contact force, such as shaking to insert a key. However, a single push is generally faster than the zig-zag push and can reduce the likelihood of the damage to the fruit. Therefore, we only use the zig-zag push in more complex situation, depending on the number and distribution of the obstacles in layer 4. 

\subsubsection{Single Push} 
\begin{figure} [ht]
    \centering
    \includegraphics[width=3in]{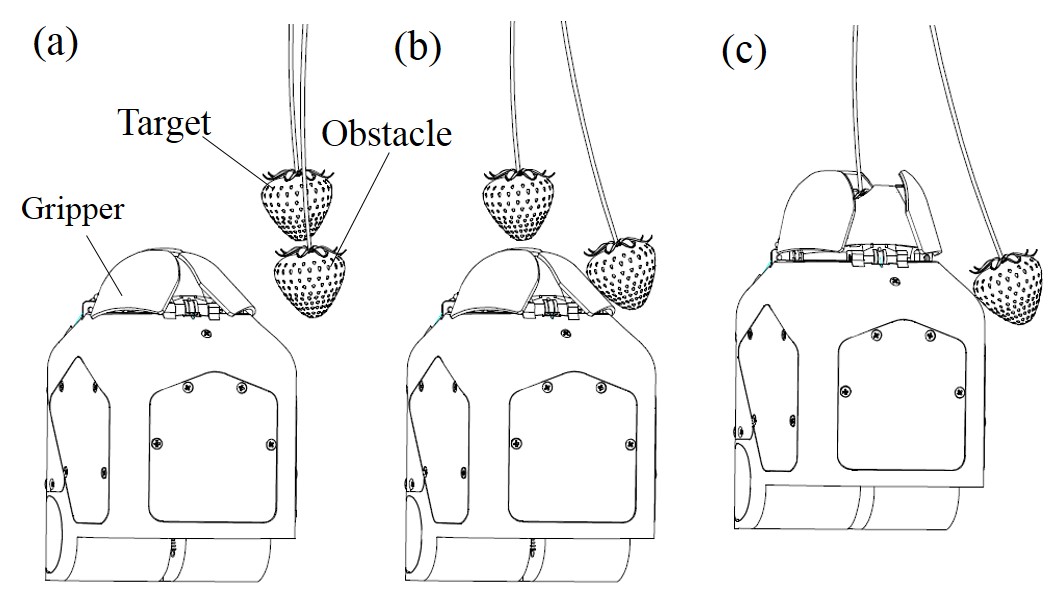}
    \caption{Singe push to move the obstacles under the target in layer 4}
    \label{fig:fig3}
\end{figure}
As shown in Fig. \ref{fig:fig3}(a), if an obstacle is located below the target (layer 4), the gripper may capture the obstacles if it moves up straightly to enclose the target. In this case, the gripper can use a single push operation to push aside the obstacle (to the right in the figure) before swallowing the target (Fig. \ref{fig:fig3}(b) and (c)). 
\begin{figure} [ht]
    \centering
    \includegraphics[width=\linewidth]{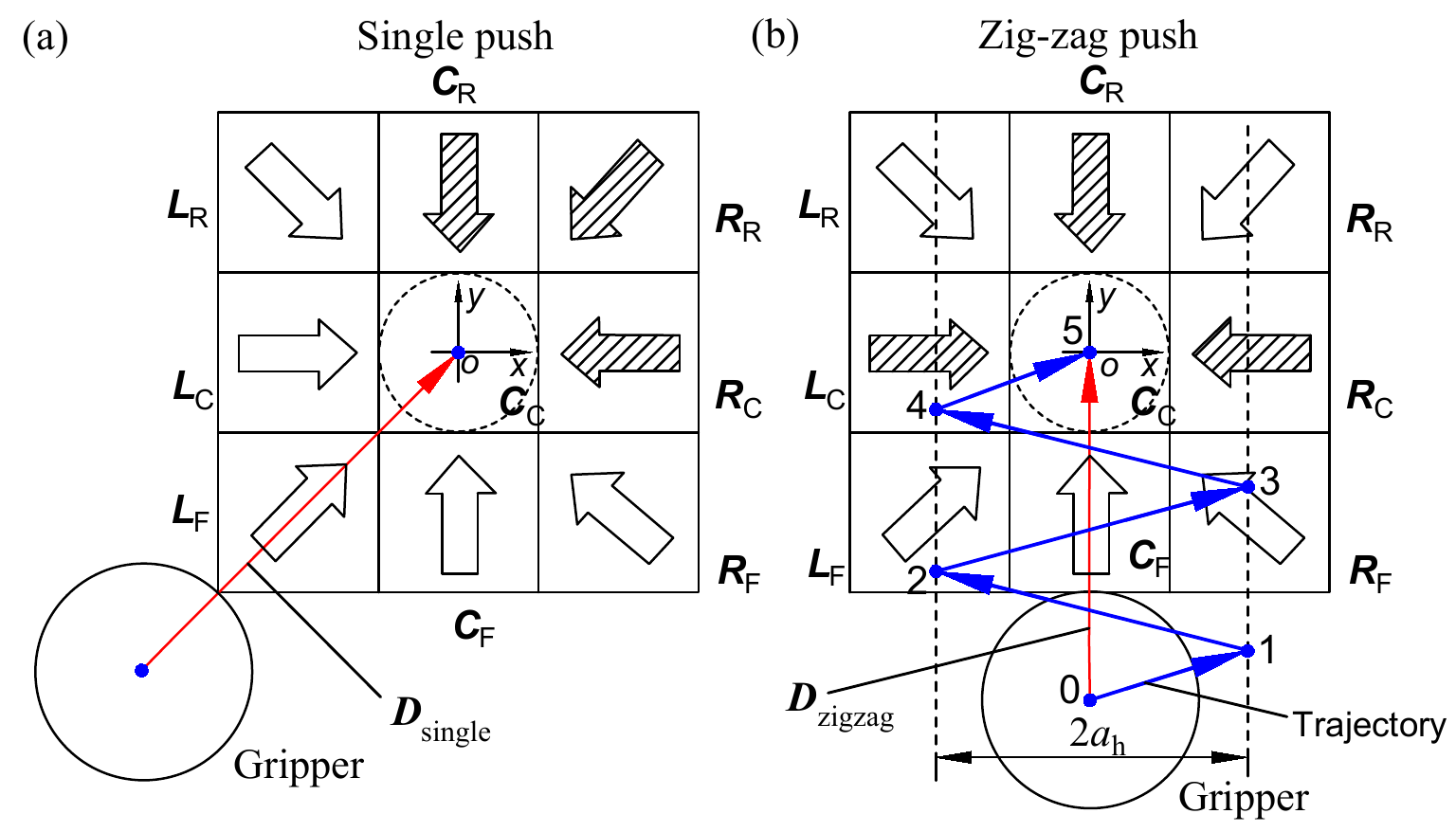}
    \caption{Diagram of the calculation of horizontal push: (a) single push, where the red arrow is the pushing direction; (b) zig-zag push, where the red arrow shows the overall direction and the blue arrows are the zig-zag paths.}
    \label{fig:fig4_3}
\end{figure}

Since the gripper size is limited, a single push operation makes it easier to move a few number of adjacent obstacles out of the way, but hard to separate sparsely distributed obstacles. Therefore, ignoring the central block, we use the number of blocks $n_{h}$ within the largest group of adjacent unoccupied blocks (without obstacles) to determine whether to use a single push or zig-zag operation. As shown in Fig. \ref{fig:fig4_3}(a), ignoring the central block, filled arrows in the blocks mean the blocks are occupied with obstacles, while the blank arrows represent unoccupied blocks. In this case, the $n_{h}$ is 5 and greater than a predetermined threshold $t_{h}$ (using 4 in this paper), so a single push operation is appropriate to push the obstacles aside. As the pushing operation is moving towards the obstacles, the direction of the single push operation for the gripper is calculated based on the positions of the occupied blocks according to the following equation:
\begin{equation} 
 \textit{\textbf{D}}_{\mathrm{single}}= -r\Sigma_{1}^n\textit{\textbf{O}}_{i}/|\Sigma_{1}^n\textit{\textbf{O}}_{i}|, \quad n_{h}>=t_{h}\\  
\label{eq:eq1}
\end{equation}
where, $\textit{\textbf{O}}_{i}$ is the vector of the $i^{th}$ occupied block within the largest group of adjacent occupied blocks and $n$ is the total number of blocks within the largest group of adjacent occupied blocks. The parameter $r$ is used to scale the $\textit{\textbf{D}}_{\mathrm{single}}$ norm, which should guarantee that the gripper pushes from the outside of the blocks (50 mm is used for the current system). The red arrow in Fig. \ref{fig:fig4_3}(a) shows the calculated pushing direction for the single push operation. It can be seen that the gripper moves from the center of the unoccupied blocks to the center of the occupied blocks, such that the gripper has the highest possibility to move all the obstacles out of the way. 

\subsubsection{Zig-zag Push}

Ignoring the central block, if the number $n_{h}$ of the largest group of adjacent unoccupied blocks comprises fewer than the threshold number $t_{h}$ of blocks, the method determines that a horizontal zig-zag push operation is appropriate. Fig. \ref{fig:fig4_3}(b) shows a case where a zig-zag operation is selected to push the obstacles side to side. The red arrow is the overall direction of the operation, while the blue arrows are the zig-zag paths. Since the zig-zag operation involves three directions of movement (forward, left and right), the gripper can push the three directions of obstacles out of the way. Therefore, it makes sense that the gripper moves from the entrance which contains fewer obstacles. Hence, different from the single push, the overall direction of the zig-zag push operation is calculated based on the positions of the unoccupied blocks according to the following equation:
\begin{equation} 
 \textit{\textbf{D}}_{\mathrm{zigzag}}= r\Sigma_{1}^m\textit{\textbf{U}}_{j}/|\Sigma_{1}^m\textit{\textbf{U}}_{j}|, \quad n_{h}<t_{h}\\  
\label{eq:eq2}
\end{equation}
where, $\textit{\textbf{U}}_{j}$ is the vector of the $j^{th}$ unoccupied block within the largest group of adjacent unoccupied blocks and $m$ is the total number of blocks within the largest group of adjacent unoccupied blocks. The norm of $\textit{\textbf{D}}_{\mathrm{zigzag}}$ is scaled using the same parameter $r$ as Eq. \ref{eq:eq1}. 
During a horizontal zig-zag push operation, the device moves in the $xy$ plane, wherein the resultant vector of the zig-zag motion is equal to $\textit{\textbf{D}}_{\mathrm{zigzag}}$ and the amplitude $a_{h}$ and number of pushes $n_{hp}$ of the zig-zag motion are determined according to the specific picking scenario. For example, the effectiveness of the values may be affected by the peduncle length, fruit weight or the damping ratio of the fruit, which are difficult to calculate. Based on some tests in the farm, in the current system, we tune the $a_{h}$ and $n_{hp}$ to fix values of 20 mm and 5, respectively. 


\subsection{In-hand Drag Operation Above The Target} 
\begin{figure} [ht]
    \centering
    \includegraphics[width=\linewidth]{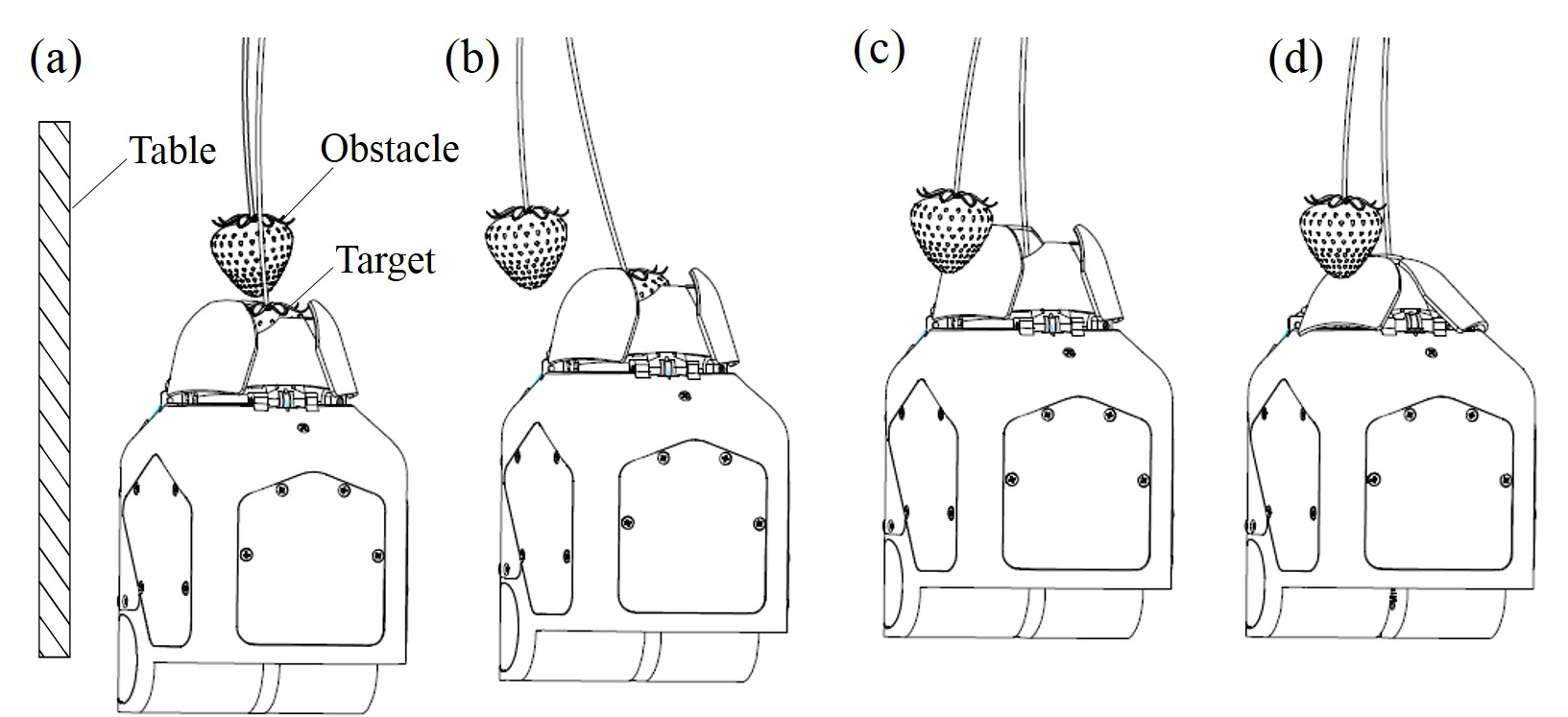}
    \caption{Drag operation to avoid capture the obstacles: an upward drag step moves the target to an area that contains fewer obstacles ((a) and (b)); an upward push-back step pushes the upper obstacles aside (c) before closing the fingers (d).}
    \label{fig:fig5}
\end{figure}

If an obstacle is located above the target (layer 1), such as the case shown in Fig. \ref{fig:fig5}(a), the gripper may swallow or damage the obstacles when moving upward to capture the target strawberry. Furthermore, the obstacles may stop the fingers closing thus resulting in cutting failure of the target peduncle. To solve this problem, we propose an in-hand drag operation, which is opposite to the push operation as used in other layers. The drag operation allows the gripper to pick the target fruit without capturing unwanted obstacles. The operation comprises an upward drag step to move the target to an area that contains fewer obstacles (Fig. \ref{fig:fig5}(b)) and an upward push-back step that pushes the upper obstacles aside (Fig. \ref{fig:fig5}(c)) before closing the fingers. The push-back step is necessary because when at the drag position (Fig. \ref{fig:fig5}(b)), the peduncle is inclined such that the fruit is difficult to fall due to the static force and easily damaged when the gripper moves up further towards a cutting position. 
\begin{figure} [ht]
    \centering
    \includegraphics[width=\linewidth]{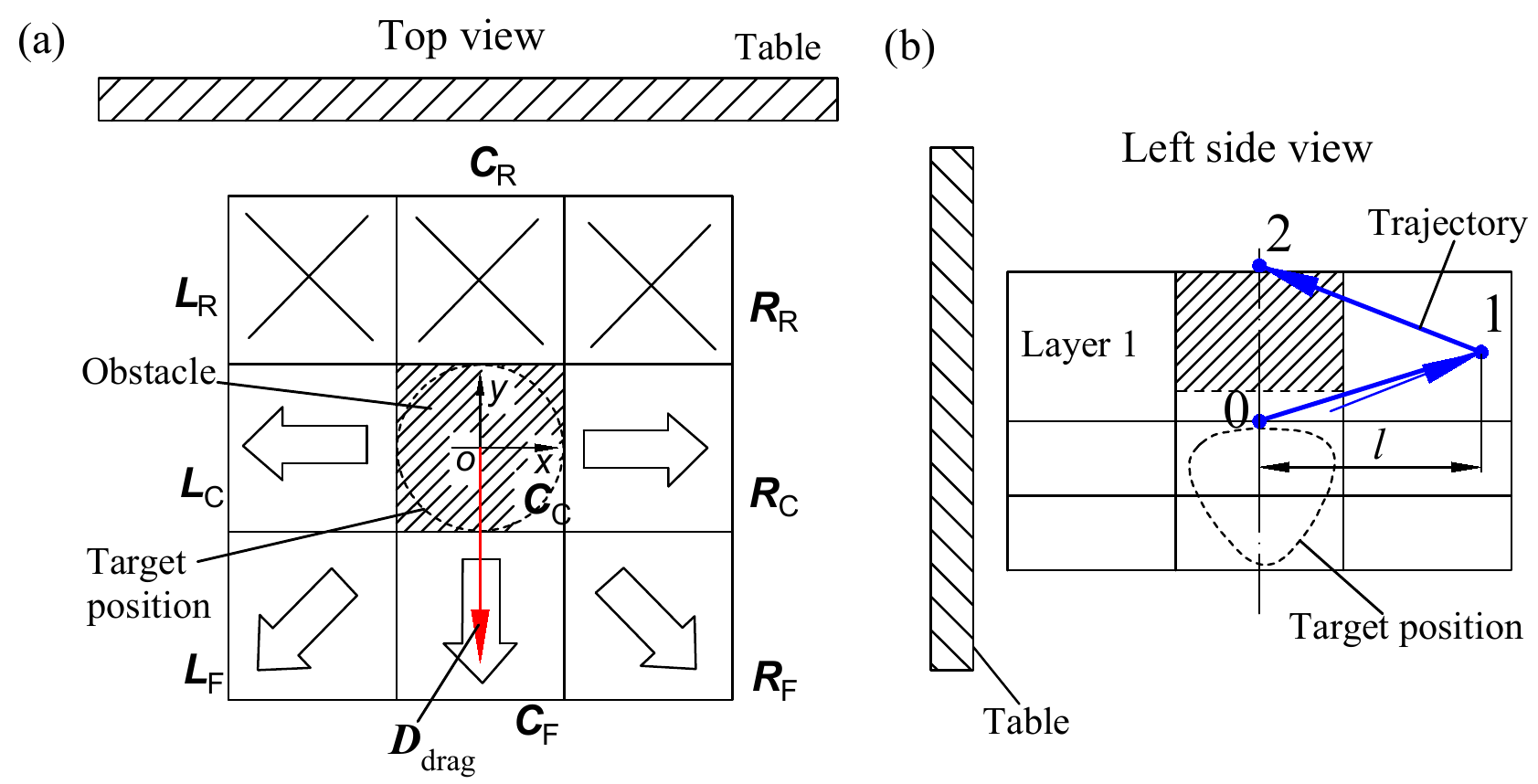}
    \caption{Diagram of the calculation of the drag operation: (a) a top view shows the calculation of the drag direction $\textit{\textbf{D}}_{\mathrm{drag}}$ in $xy$ plane; (b) a left side view shows the drag and push-back paths (blue arrows).}
    \label{fig:fig5_2}
\end{figure}

The drag operation is performed only when there are obstacles in the central block $\textit{\textbf{C}}_{\mathrm{C}}$ of the top layer. If the $\textit{\textbf{C}}_{\mathrm{C}}$ is unoccupied, the gripper moves directly upwards to pick the target strawberry. Fig. \ref{fig:fig5_2} shows the diagram of the calculation method of the drag operation with corresponding to the example in Fig. \ref{fig:fig5}. As shown in Fig. \ref{fig:fig5_2}(a), to avoid the collision between the gripper and the table, the three blocks $\textit{\textbf{L}}_{\mathrm{R}}$, $\textit{\textbf{C}}_{\mathrm{R}}$, $\textit{\textbf{R}}_{\mathrm{R}}$ that are close to the table are skipped for the calculation of the drag direction. Then the drag direction $\textit{\textbf{D}}_{\mathrm{drag}}$ in the $xy$ plane can be determined according to the following equation:
\begin{equation} 
\label{eq:eq5}
             \textit{\textbf{D}}_{\mathrm{drag}}=l\Sigma_{1}^m\textit{\textbf{U}}_{j}/|\Sigma_{1}^m\textit{\textbf{U}}_{j}|  
\end{equation}
where, $\textit{\textbf{U}}_{j}$ is the vector of the $j^{th}$ unoccupied block within the largest group of adjacent unoccupied blocks. The blocks used for calculation are $\textit{\textbf{L}}_{\mathrm{C}}$, $\textit{\textbf{L}}_{\mathrm{F}}$, $\textit{\textbf{C}}_{\mathrm{F}}$, $\textit{\textbf{R}}_{\mathrm{F}}$, $\textit{\textbf{R}}_{\mathrm{C}}$. The parameter $m$ is the total number of blocks within the largest group of adjacent unoccupied blocks. The norm of $\textit{\textbf{D}}_{\mathrm{drag}}$ is scaled to $l$ (50 mm is used in the current system). If all blocks are occupied by obstacles, the drag direction is aligned to $\textit{\textbf{C}}_{\mathrm{F}}$, where there are fewer obstacles in general. Fig. \ref{fig:fig5_2}(b) shows the drag and push-back steps, wherein the drag and push-back operations moves up the same height in the vertical direction. 

\section{EXPERIMENTS}
\subsection{Image Processing} 
\begin{figure*} [b!]
    \centering
    \includegraphics[width=\linewidth]{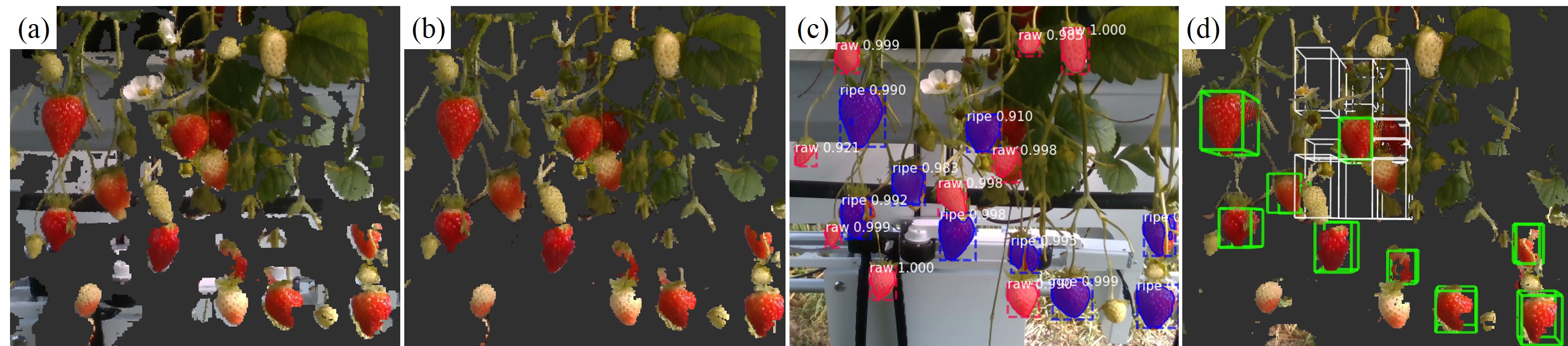}
    \caption{The workflow of the fruit detection and obstacle determination: (a) original point cloud captured by an RGB-D camera; (b) 3D HSI color thresholding to remove the adjacent noise points from the background; (c) using deep learning to detect the ripe strawberries in an RGB image; (d) fruit localization and obstacle determination in 3D point cloud.}
    \label{fig:fig7}
\end{figure*}

The image processing includes the detection and localization of ripe strawberries and also the determination of obstacles within the RoI for each target. An RGB-D camera (D435; Intel, USA) was used for image acquisition. The image processing pipeline contains three steps: 1) 3D color thresholding to remove noise points from the background, 2) object detection and localization using deep learning based on our previous work \cite{ge2019access} and 3) obstacle calculation. 

Fig. \ref{fig:fig7}(a) shows the original point cloud, where some pieces of points from the table (silver) and irrigation tubes (black) are around the strawberries. In fact, the table and irrigation tubes are behind the berries at a distance of about 150 mm. The inaccurate depth sensing results in some of the points connecting to the front berries, which may be regarded as obstacles. To avoid this influence, the first step is to remove the adjacent noise points (silver and black) by using hue, saturation and intensity (HSI) color thresholding, as the result is shown in Fig. \ref{fig:fig7}(b). This step is performed in point cloud using the \textit{jsk\_pcl\_ros} ROS package.  
The second step is the detection and localization of the ripe strawberries. As shown in Fig. \ref{fig:fig7}(c), we use an instance segmentation convolutional neural network Mask R-CNN to segment the objects in pixel level such that the 3D location of the ripe strawberries can be obtained by matching with depth images \cite{ge2019access}. The detection system outputs the 3D bounding boxes of the target strawberries and the thresholded point cloud for further obstacle calculation, as shown in Fig. \ref{fig:fig7}(d). 
The obstacles around the target is determined based on the method described in Section \ref{RoI}. To calculate the number of points in each block, we crop the bounding box of each RoI block in point cloud using the \textit{CropBox} function in the Point Cloud Library (PCL). Fig. \ref{fig:fig7}(d) shows the obstacle bounding boxes (white) around a target, where only blocks occupied with obstacles are displayed.

\subsection{Field Test Setup} 
\begin{figure} [ht]
    \centering
    \includegraphics[width=2.3in]{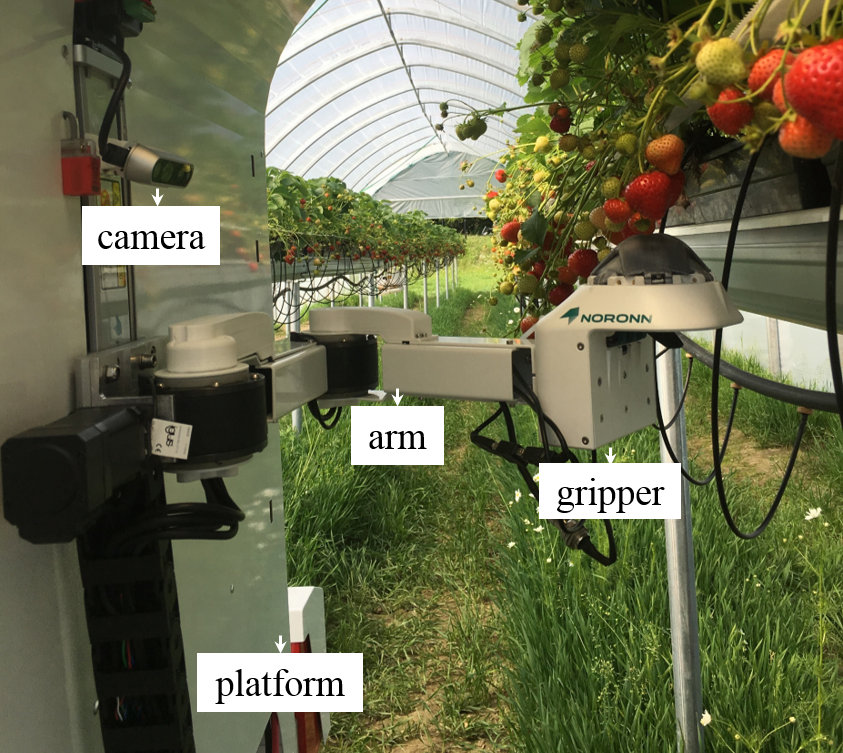}
    \caption{The newly developed U-shaped strawberry-harvesting robot on a farm: inner view of the picking system (NORONN, \url{www.noronn.com}).}
    \label{fig:fig6}
\end{figure}

We conducted two sets of experiments in two places: a greenhouse at the Boxford Suffolk Farms (England) and a university experimental tunnel at the Norwegian University of Life Sciences (Norway). The tests were carried out on strawberry cultivars of ``Malling Centenary" in England and ``Murano" in Norway. Generally, ``Malling Centenary" is easier for obstacle separation because most of the berries have long and independent peduncles, while ``Murano" berries have more clusters with short peduncles growing on one stem. The different biological characteristics may result in different performance of the robot. The field tests were performed on our newly developed U-shaped strawberry-harvesting robot with a newly developed SCARA-like arm, as shown in Fig. \ref{fig:fig6}.

\begin{figure*} [b!]
    \centering
    \includegraphics[width=\linewidth]{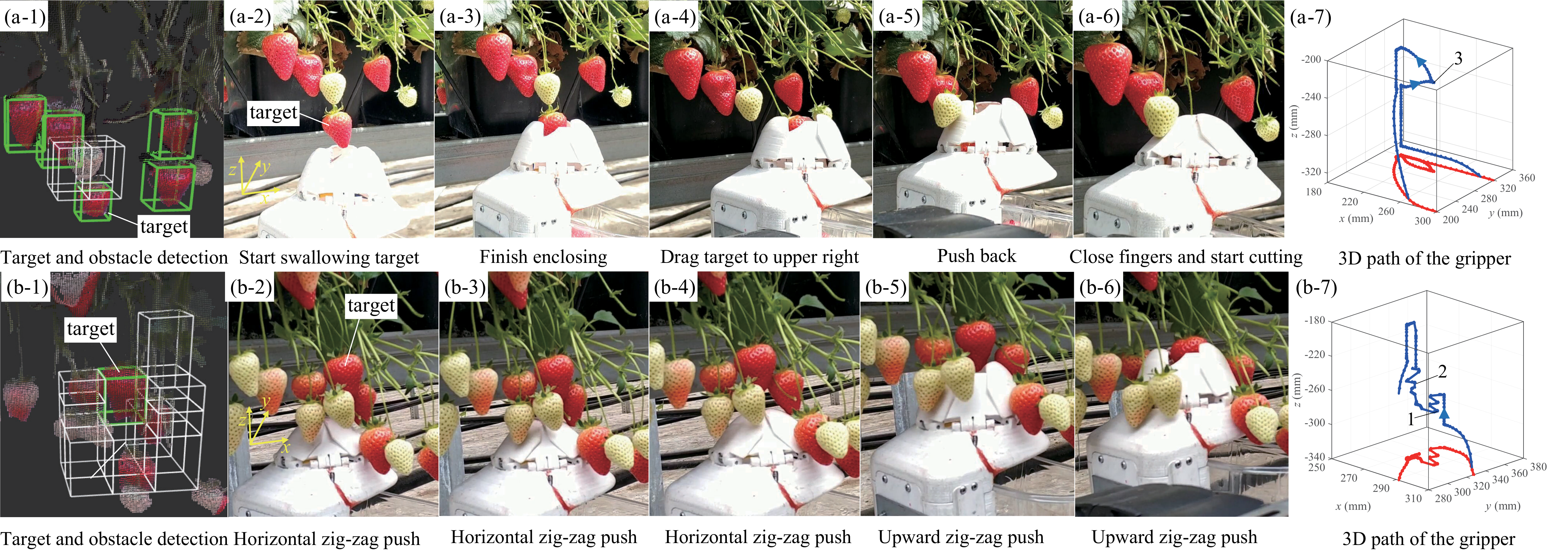}
    \caption{Examples of the obstacle separation in the field test: each row of images represents a picking case, where the first (left) image shows the detection and localization of target and obstacles and the last image (right) displays the 3D path of the gripper.}
    \label{fig:fig8}
\end{figure*}

\subsection{Results} 

\subsubsection{Application examples}

Fig. \ref{fig:fig8} demonstrates two examples of using the proposed obstacle separation method in different situations. In Fig. \ref{fig:fig8}(a), a green obstacle is located above the target, which may be mis-captured when the gripper is open to capture the target. Fig. \ref{fig:fig8}(a-1) shows that there are no obstacles in layer 2, 3 and 4, but in layer 1, the central block $\textit{\textbf{C}}_{\mathrm{C}}$ and three other blocks $\textit{\textbf{L}}_{\mathrm{C}}$, $\textit{\textbf{L}}_{\mathrm{F}}$ and $\textit{\textbf{C}}_{\mathrm{F}}$ are occupied with obstacles, so based on Eq. \ref{eq:eq5}, a drag operation is required. As there are no obstacles in layer 2, 3 and 4, the gripper moves up directly to enclose the target, as shown in Fig. \ref{fig:fig8}(a-2) and (a-3). After holding the target, the gripper drags the target to the front-right region while moving upward where it contains fewer obstacles (Fig. \ref{fig:fig8}(a-4)). At this position, if the gripper continues to move up, it might be difficult for the target to fall down towards a cutting position because the peduncle is inclined and the target has a contact force with the fingers. It may also damage the target with such a force. Therefore, in Fig. \ref{fig:fig8}(a-5), the gripper pushes back to the central position while moving up for further fruit detachment (Fig. \ref{fig:fig8}(a-6)), in which the upper obstacles are pushed aside. The blue line in Fig. \ref{fig:fig8}(a-7) shows the recorded 3D trajectory of the gripper during the operation, while the red line is the trajectory projection on the $xy$ plane, from which we can see that the gripper drags the target to $-y$ and $+x$ direction and then moves back. In the last image of each case, paths 1, 2 and 3 represent the three stages of operations in the bottom layer, central layers and top layer, respectively. 

Fig. \ref{fig:fig8}(b) demonstrates one more examples of the obstacle separation algorithm with more obstacles in the central layers and bottom layer. Only two blocks ($\textit{\textbf{L}}_{\mathrm{R}}$ and $\textit{\textbf{C}}_{\mathrm{R}}$) in layer 4 are unoccupied with obstacles for the target in Fig. \ref{fig:fig8}(b-1). Therefore, the gripper uses the horizontal zig-zag push from the left rear to right front (the red line in Fig. \ref{fig:fig8}(b-7) shows the direction) to push the obstacles side to side, as can be seen in Fig. \ref{fig:fig8}(b-2) to (b-4). Then, the gripper continues to use an upward zig-zag push operation to separate the obstacles in the central layers (Fig. \ref{fig:fig8}(b-5) and (b-6)). Without this operation, the target and the obstacles may not be separated but pushed up together due to the contact force between each other.

\subsubsection{Performance test and failure cases} 

We recorded the test data in England and Norway under different settings to analyze the feasibility and performance of the obstacle separation method. The tests in England only show the results using the obstacle separation method, while in Norway, we conducted a comparison test, using and without using the obstacle separation method. For each setting, we implemented 100 attempts for the detected targets. Also, as the focus in this study is the obstacle separation method, only the targets with obstacles were used for the tests. This may result in a lower success rate compared to our previous reports, because the robot has a good performance on isolated strawberries \cite{xiong2018design}. 
\begin{figure} [ht]
    \centering
    \includegraphics[width=3in]{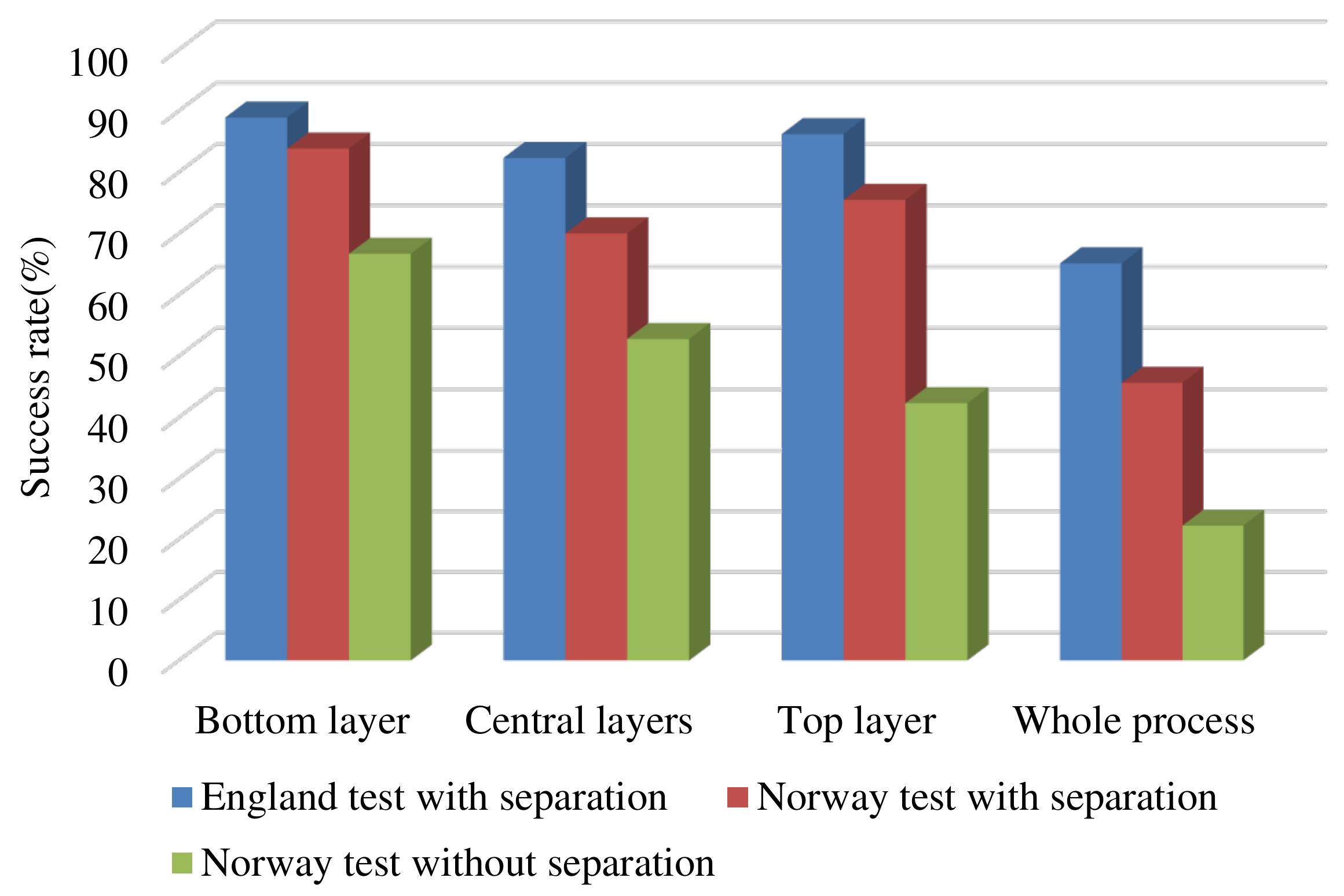}
    \caption{Comparison of success rates in different layers under different settings.}
    \label{fig:fig9}
\end{figure}

Fig. \ref{fig:fig9} reports the success rates in each stage and also the whole process under different settings. The whole process means the manipulation in the whole three stages in the bottom layer, central layers and top layer. In each independent stage, we only considered the results with obstacles in the corresponding layers, while a whole process may contain zero obstacle in one or two stages but at least one obstacle in all the layers. The success in the whole process means that in all the stages the separation is successful. Generally, the comparison tests in Norway show that the obstacle separation method is effective compared to the attempts without using the obstacle separation method. Also, the variety of ``Malling Centenary" tends to be easier to be picked compared to ``Murano". To be more precise, the comparison tests show that the drag operation in the top layer is most effective, increasing the success rate from 42.3\% to 75.5\% in the Norway tests. The bottom layer is the easiest layer in terms of obstacle separation. This might be attributed to the gripper design, since the opening action of the fingers under the target can help to push the obstacles aside. The success rate of the whole process is relatively low compared to the operation in the individual layer. For the test on ``Murano", the success rate of the whole process increases from 22.2\% to 45.6\% by using the separation method. The same separation method shows a better performance (65.1\%) on the variety of ``Malling Centenary".

\section{CONCLUSIONS}
We present an active obstacle separation method for selectively picking a target fruit surrounded by obstacles. Different from the traditional obstacle avoidance methods, the proposed separation method combining push and drag motions to actively separate obstacles from the target based on 3D visual perception. In addition to the old single linear push, a zig-zag push operation that contains several linear pushes was used for both bottom layer and central layers of the target, which is able to separate more dense obstacles. Moreover, the generated side-to-side motion can break the static contact force between the target and obstacles, thus making it easier for the gripper to receive the target. Furthermore, we proposed a novel drag operation to address the issue of mis-capturing obstacles located above the target, in which the gripper drags the target to a place with fewer obstacles and then pushes back to move the obstacles aside for further detachment. The separation paths are calculated based on the number and distribution of the obstacles. 
Also, an image processing pipeline was developed to implement the method on a harvesting robot. Field tests showed that the proposed method can improve the picking performance substantially. The performance may be further improved by incorporating a closed-loop vision guided manipulation system. A video of the field experiments can be found at  \url{https://drive.google.com/file/d/15BO2_4aaR5KHxbgOJQ9fV76zf0_i4GJs/view?usp=sharing}.






\bibliographystyle{IEEEtran}
\bibliography{IEEEexample}

\end{document}